\pdfoutput=1

\documentclass[11pt]{article}

\usepackage{acl}

\usepackage{times}
\usepackage{latexsym}
\usepackage[T1]{fontenc}
\usepackage[utf8]{inputenc}
\usepackage{microtype}
\usepackage{times}
\usepackage{latexsym}
\usepackage{graphicx}
\usepackage{todonotes}
\usepackage{csquotes}
\usepackage{dirtytalk}
\usepackage{hyperref}
\usepackage{booktabs}
\usepackage{enumitem}
\usepackage{multirow}
\usepackage{subcaption}
\usepackage{xurl}

\title{Towards Lexical Gender Inference: A Scalable Methodology using Online Databases}

\author{Marion Bartl \\
  Insight SFI Research Centre \\
  for Data Analytics\\
  School of ICS\\
  University College Dublin\\
  \texttt{marion.bartl@insight-centre.org} \\ \And
  Susan Leavy \\
  Insight SFI Research Centre \\ 
  for Data Analytics \\
  School of ICS\\
  University College Dublin\\
  \texttt{susan.leavy@ucd.ie} \\}

\begin{document}
\maketitle
\begin{abstract}

This paper presents a new method for automatically detecting words with lexical gender in large-scale language datasets. Currently, the evaluation of gender bias in natural language processing relies on manually compiled lexicons of gendered expressions, such as pronouns (\textit{he}, \textit{she}, etc.) and nouns with lexical gender (\textit{mother}, \textit{boyfriend}, \textit{policewoman}, etc.). However, manual compilation of such lists can lead to static information if they are not periodically updated and often involve value judgements by individual annotators and researchers. Moreover, terms not included in the list fall out of the range of analysis. 
To address these issues, we devised a scalable, dictionary-based method to automatically detect lexical gender that can provide a dynamic, up-to-date analysis with high coverage. Our approach reaches over 80\% accuracy in determining the lexical gender of nouns retrieved randomly from a Wikipedia sample and when testing on a list of gendered words used in previous research.

\end{abstract}

\section{Introduction}
There is a growing body of research on gender bias embedded in trained language models as well as on allocational and representational harms caused by the deployment of these models. There have moreover been increasing calls for early and thorough data description and curation in order to gain insights into how, for instance, gender stereotyping or quality of service bias is propagated from data into a language model. What both of these strands of research have in common is their reliance on pre-defined lexicons of terms related to gender.

In English, gendered words most commonly include pronouns (\textit{he}, \textit{she}, \textit{they}, etc.), and also words that carry lexical gender, such as \textit{boyfriend}, \textit{policewoman}, or \textit{prince}. Previous works on gender bias in language technologies often use manually compiled lists of words carrying lexical gender to, for example, mitigate gender stereotyping through data augmentation \citep{lu2020CDA}, assess trans-exclusionary bias in co-reference annotations \citep{cao2020toward} or evaluate gender inequalities in Wikipedia article titles \citep{falenska_assessing_2021}. However, curated lists are limited in their coverage of terms that contain lexical gender and can become outdated if not maintained. 

To address this issue, we present a scalable algorithmic method to determine lexical gender by querying a word's dictionary definitions for a small subset of definitively gendered words. Our method allows for high-coverage, instantaneous detection of words carrying lexical gender, which eliminates the need to manually compile and maintain static lists of gendered words. This not only facilitates the extension of previous work on gender bias in NLP, but can also be used for a more detailed analysis on the representation of gender in large-scale language datasets used to train large language models like BERT \citep{devlin-etal-2019-bert} or GPT-2 \citep{radford2019language}.

By combining the gender labels obtained from Merriam Webster Online \citep{merriam_webster}, WordNet\textsuperscript {\textregistered} \citep{wordnet} and Dictionary.com \citep{dictionarycom_2022}, our method reaches an accuracy of 84\% in determining the lexical gender of words in a random sample of 1,000 Wikipedia articles and 87\% accuracy on a list of words carrying lexical gender adapted from previous research. The code for the algorithm, evaluation methods and datasets are available\footnote{\url{https://github.com/marionbartl/lexical-gender}}.

In the following section we first outline the conceptions of linguistic gender used in this research and secondly present an overview of research on gender in language technology that relies on curated lists of gendered words. Thirdly, we discuss prior approaches to algorithmic gender inference. Section \ref{sec:methods} gives a detailed overview of the algorithm and Section \ref{sec:data} introduces the datasets used to assess our gender detection algorithm. We present quantitative and qualitative results in Section \ref{sec:results} and discuss limitations as well as avenues for future development. 

\section{Background}
When dealing with the category of gender in language technology, it is important to make a distinction between the social category of gender and gender in a linguistic sense. While social gender relates to the complex property, performance and experience of one's own and others' gender within society \citep{ackerman2019syntactic}, linguistic gender describes the expression of gender within grammar and language. 
In English, linguistic gender mainly encompasses ways to express gender as female, male or gender-indefinite \citep{fuertes-olivera2007lexical}.
Social gender, as an extra-linguistic category, includes a more fluid view of gender aside from male and female categories. This includes transgender, genderqueer and other non-binary experiences and expressions of gender \citep{darwin2017doing}.
As \citet{bucholtz1999gender} and \citet{cao2020toward} point out, there is no ``one-to-one'' mapping between social and linguistic gender. 
However, the two are influenced by each other: on one hand, expressions of gender in language are subject to changing norms in society \citep{fuertes-olivera2007lexical}, on the other hand, the way gender is represented in language influences the conception of gender within society \citep{butler_gender_1990}. Thus, being able to evaluate gendered expressions in language provides insights into societal conceptualisations of gender.

Since this research explicitly focuses on lexical gender in English, which is a linguistic category, we give an overview of linguistic gender in English in Section \ref{ssec:ling_gender}. 
Section \ref{ssec:NLP_background} explores the role lexical gender information plays in different areas of research on gender bias in NLP, which simultaneously present possible areas of application for our method of lexical gender inference. 
Section \ref{ssec:inference} discusses two prior algorithmic systems for lexical gender inference in English.

\subsection{Linguistic gender in English}\label{ssec:ling_gender}

The taxonomy of linguistic gender in this work builds upon the approach developed by \citet{cao2020toward} and incorporates work by \citet{corbett1991gender}, \citet{hellingerbuss2003gender} and \citet{fuertes-olivera2007lexical}. 

Within linguistic gender, \citet{cao2020toward} differentiate between grammatical, referential, and lexical gender. \textbf{Grammatical gender} refers to the distinction of noun classes based on agreement between nouns and their dependants. English, as a natural or notional gender language \citep{McConnellGinet+2013+3+38}, does not have grammatical gender, but it has referential and lexical gender. 
\textbf{Referential gender} is used to refer to the social gender of a specified extra-linguistic entity. Thus, it ``relates linguistic expressions to extra-linguistic reality, typically identifying referents as `female', `male', or `gender-indefinite.'~'' \citep{cao2020toward}. In English, pronouns fall under the category of referential gender. 
\textbf{Lexical gender}, which this work focuses on, is non-referential but a semantic property of a given linguistic unit, which can be either masculine, feminine\footnote{We use the terms \textit{masculine} and \textit{feminine} instead of \textit{male} and \textit{female} here in order to underline the purely linguistic, i.e. semantic, property of lexical gender} or gender-indefinite/gender-neutral. \citet{ackerman2019syntactic} calls these words ``definitionally gendered''. 
Words that carry lexical gender can require semantic agreement in related forms, such as, for instance, using the pronoun \textit{his} in connection with the word \textit{stuntman} in the sentence `Every \textit{stuntman} needs to rehearse \textit{his} stunts.' \cite{fuertes-olivera2007lexical}. 
In English, lexical gender is usually not morphologically marked. Exceptions to this rule include e.g. the suffixes \textit{-man} to denote masculine gender, such as in \textit{policeman}, or \textit{-ess} to denote feminine gender, such as in \textit{waitress}. It should moreover be noted that lexical gender is exclusively a linguistic property. However, words containing lexical gender can be used to express referential gender if a concrete referent is specified \citep{cao2020toward}.

\subsection{Lexical gender in gender bias research}
\label{ssec:NLP_background}

The evaluation and mitigation of gender biases in language datasets and models relies on referential expressions of gender, such as pronouns, but also words that carry lexical gender. These pieces of research vary in application, as well as the number of gendered expressions considered, which varies from two to around 120 words. Most works assess binary differences between male and female gender. However, an emergent strand of NLP research also focuses on non-binary gender expressions \citep{cao2020toward} and creating gender-neutral datasets and systems \citep{vanmassenhove-etal-2021-neutral}. The following considers example use-cases of lexicons of lexically gendered words. These simultaneously represent a variety of applications for our lexical gender detection algorithm. 

\paragraph{Dataset evaluation}
The most straightforward form of using gendered words is to assess the distribution of gendered words in a corpus. \citet{zhao-etal-2019-gender} counted \textit{he}/\textit{she} pronouns in the One Billion Word Benchmark \citep{chelba2013billion} to show male skew in the training data for the ELMo language model \citep{peters-etal-2018-deep}, which is the primary focus of their analysis. This analysis addressed calls for better data evaluation \citep{bender2021dangers, rogers_changing_2021} prior to or alongside model bias analyses.

\paragraph{Retrieval for analysis}
Limited-scope lists of words that carry lexical gender were used by \citet{caliskan_semantics_2017} to retrieve Word2Vec embeddings \citep{mikolov2013efficient} and perform the Word Embedding Association Test (WEAT). This test measured stereotyping by calculating implicit associations between eight male/female word pairs and words related to maths or science and arts. \citet{guo2021detecting} used an adapted version of the WEAT, the CEAT, to asses intersectional biases in contextualized word embeddings (ELMo \citep{peters-etal-2018-deep}, BERT \citep{devlin-etal-2019-bert}, OpenAI GPT \citep{radford2019language, brown2020language}).
Another use-case in which gendered words were used for retrieval is research by \citet{falenska_assessing_2021}, who assessed gender bias in Wikipedia articles. As a first step, they filtered the article titles for a limited number of words that carry lexical gender.

\paragraph{Creation of synthetic evaluation data}
In sentence-based analyses of gender-bias, lists of words with lexical gender can also be used to fill placeholders in sentence templates and thus create synthetic sentences with different gendered entities. For example, \citet{kiritchenko-mohammad-2018-examining} created the Equity Evaluation Corpus (EEC) to analyse gender stereotyping in sentiment analysis systems. The EEC inspired the creation of the Bias Evaluation Corpus with Professions (BEC-Pro), which was used to analyse associations between gendered entities and professions in BERT \citep{bartl-etal-2020-unmasking}.
Similarly, \citet{sheng-etal-2019-woman} used the word pair \textit{the man}/\textit{the woman} as fillers within sentence-start prompts for open-ended natural language generation (NLG) and the subsequent analysis of gender biases in the generated sentences.

In a rare instance of research on non-binary representations of gender in NLP, \citet{cao2020toward} used gendered lists of words to find and hide lexical gender in the GAP dataset \citep{webster2018GAP}. The dataset created in this way was used to measure gender- and trans-exclusionary biases in coreference resolution performed by both humans and machine-learning models. 

\paragraph{Data manipulation}
Extensive lists of gendered words were used in the context of Counterfactual Data Augmentation (CDA), which replaces words with masculine lexical gender with their feminine variants and vice versa in a corpus. This is done in order to create training or fine-tuning data for gender bias mitigation.
For instance, \citet{lu2020CDA} `hand-picked' gender pairs to swap in CDA and \citet{maudslay2019CDS} added first names to the list of words to be swapped.

Another kind of data manipulation, this time aiming 
for neutral gender, was performed by \citet{vanmassenhove-etal-2021-neutral}. They used lists of unnecessarily gendered job titles (e.g. \textit{mailman/mailwoman}) and 
feminine forms (e.g. \textit{actress}), as well as generic uses of the suffix \textit{-man} (such as in \textit{freshman}) in the extended version of their \textit{Neutral Rewriter}, which re-writes explicit mentions of gender into their gender-neutral variants (\textit{mail carrier}, \textit{actor}, \textit{first-year student}). 

\subsection{Lexical gender inference}
\label{ssec:inference}
Previous approaches to automatic lexical gender inference used unsupervised and semi-supervised learning, drawing on the presence of gendered pronouns in the context of a given noun \citep{bergsma2006bootstrapping, bergsma2009glen}. While \citet{bergsma2006bootstrapping} created a large dataset of probabilistic noun gender labels, \citet{bergsma2009glen} used these as basis for creating training examples for a statistical model that uses context and morphological features to infer lexical gender. 

One major point of criticism here lies in the probabilistic determination of noun gender, which has the risk of mislabelling lexically neutral nouns, such as professions, as being gendered due to contextual distributions that are representative of stereotypes or the number of men and women holding the profession instead of the linguistic category of lexical gender.
For example, since there are more female than male nurses \citep{bureau_of_labor_statistics_bls_labor_2022} and thus most nurses are referred to with female pronouns in text, the algorithm might infer that the term \textit{nurse} has female lexical gender, when in fact it is neutral. 


\begin{figure*}
    \centering
    \includegraphics[width=\textwidth]{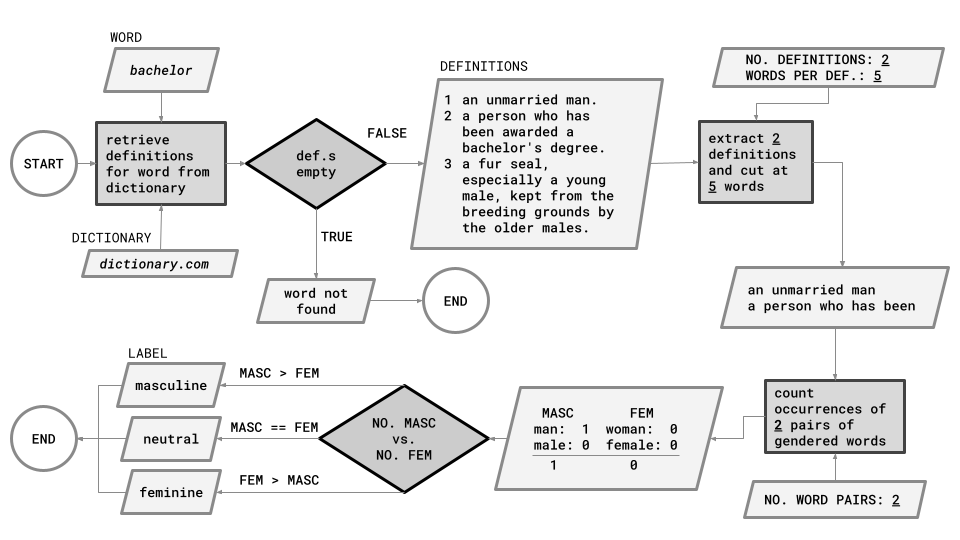}
    \caption{Simplified exemplary flowchart of gender detection algorithm}
    \label{fig:flowchart}
\end{figure*}

\section{Method: Automatic Detection of Lexical Gender}\label{sec:methods}

The main goal of this work is to produce a dynamic, high coverage, scalable method to determine the lexical gender of a target word in order to replace previously used manually compiled lexicons. For this purpose, we leveraged the fact that the definition of a lexically gendered word includes words from a small set of definitively gendered words that carry the same lexical gender. In the following, we describe the main algorithm setup, additional parameters and heuristics, as well as the method to combine lexical gender labels from different databases. A schematic, exemplary overview of the algorithm is presented in Figure \ref{fig:flowchart}.

\begin{table*}[t]
\renewcommand\arraystretch{1.2}
\centering
\resizebox{0.8\textwidth}{!}{%
\begin{tabular}{@{}ccccccccc@{}}
\toprule
\multicolumn{1}{l}{\textbf{$w$}} & 1     & 2      & 3       & 4        & 5      & \multicolumn{1}{l}{6} & \multicolumn{1}{l}{7} & \multicolumn{1}{l}{8} \\ \midrule
\textbf{feminine}                & woman & female & wife    & daughter & mother & girl                  & sister                & aunt                  \\
\textbf{masculine}               & man   & male   & husband & son      & father & boy                   & brother               & uncle                 \\ \bottomrule
\end{tabular}%
}
\caption{Words carrying explicit lexical gender; $w$ = number of pairs used for experiments}
\label{tab:seed_words}
\end{table*}

\subsection{Algorithm construction}
The method we outline utilises the increasing availability of machine-readable dictionaries, such as Merriam Webster Online, Dictionary.com, and the lexical database WordNet, in order to identify gendered terms. Examples \ref{example:nun} and \ref{example:monk} illustrate how lexical gender is captured within Merriam-Webster's \citeyearpar{merriam_webster} definitions of \textit{nun} and \textit{monk}:

\begin{enumerate}[label={(\arabic*})]
\item \textit{nun}\label{example:nun}: a woman belonging to a religious order
\item \textit{monk}\label{example:monk}: a man who is a member of a religious order and lives in a monastery
\end{enumerate}

Both definitions mention the lexical gender of the referent through a gendered word, in this case \textit{man} and \textit{woman}. Initial analyses showed that gendered words are more likely to occur at the beginning of a definition and definitions often used the words \textit{female/male} or \textit{woman/man} to specify lexical gender. In identifying gendered terms, we thus considered the presence and amount of up to eight definitively gendered words, such as \textit{male/female}, \textit{man/woman} etc., in the target word's definitions to draw inferences about its lexical gender.

For retrieval of the definitions, we accessed WordNet through the Natural Language Toolkit (NLTK,  \citeauthor{bird2009nltk}, \citeyear{bird2009nltk}) and Merriam Webster Online as well as Dictionary.com through HTTP requests. 

Once the definitions for a given target word were retrieved, the process of obtaining lexical gender was the same for either dictionary.
We determined whether a word has masculine, feminine, or neutral lexical gender by counting occurrences of a number of word pairs which have clearly defined feminine or masculine lexical gender, which are displayed in Table \ref{tab:seed_words}. If the combined definition texts contain more masculine than feminine terms, the word was labelled with masculine lexical gender, and vice versa. If the same number of masculine and feminine words was found within a set of definitions, which includes the case in which none of the pre-defined gendered terms can be found, the word was labelled with neutral lexical gender.
We additionally obtained a combined label through a majority vote over the individual dictionaries' gender labels. In cases in which words could not be found in one dictionary and querying each of the other dictionaries returned different labels, a neutral gender label was assigned.

\subsection{Parameters}

Three variable parameters were used to limit the number of definitions and word tokens queried, as well as the number of definitively gendered words to use for the query. In order to determine the best combination of values for our parameters, we performed a grid search using our gold standard data (see Section \ref{ssec:gold_standard}) and combined labels to test performance.

\paragraph{Number of definitions $d$}
We limited the number of definitions, because definitions that occur early on have a higher likelihood of describing a more general sense of the word, while later definitions relate to very specific word senses. Therefore, we retrieved only the first $d$ definitions that the dictionary lists for the word.
During grid search, we tested integer values in the range $d=[2..10]$, and the best value was determined to be $d=4$. 

\paragraph{Number of tokens $t$}
We also experimented with limiting the number of tokens within a given definition to see whether definitively gendered terms were more likely to be mentioned earlier in a given definition. The definitions were tokenized using NLTK \citep{bird2009nltk}. We took the first $t$ tokens of each definition. Regarding the number of tokens in a definition, we tested the algorithm with $t=\{5, 10, 15, 20, 25, 30, 35\}$ in our experiments and found $t=20$ to produce optimal results.

\paragraph{Number of gendered word pairs $w$}
The word pairs used during experiments are listed in Table \ref{tab:seed_words}. 
The first two word pairs, \textit{woman/man} and \textit{female/male}, as well as the pair \textit{girl/boy}, are most commonly used to describe the gender of a person or animal, while the rest of the words describes gendered family relations. The latter were chosen in order to account for cases in which the lexical gender of a person is described in relation to another person by using family terms. This is for example the case in the definition of \textit{baroness} in Merriam Webster: ``the wife or widow of a baron'' \citep{merriam_webster}. 
The grid search was performed for integer values in the range $w=[2..8]$ and best performance was obtained for $w=5$ word pairs. 
Moreover, if a target word was included in the definitively gendered pairs or their plural forms, it was automatically classified with the respective lexical gender.

\begin{table*}[ht]
\centering
\begin{tabular}{@{}lccccccc@{}}
\toprule
 &
  \textbf{\begin{tabular}[c]{@{}c@{}}gold\\ (N=134)\end{tabular}} &
  \multicolumn{3}{c}{\textbf{\begin{tabular}[c]{@{}c@{}}Wiki1000-sample\\ (N=515)\end{tabular}}} &
  \multicolumn{3}{c}{\textbf{\begin{tabular}[c]{@{}c@{}}Wiki1000 dataset\\ (N=12,643)\end{tabular}}} \\ \midrule
\textbf{POS} &
  \textbf{NN} &
  \multicolumn{1}{l}{\textbf{NN}} &
  \multicolumn{1}{l}{\textbf{NNS}} &
  \multicolumn{1}{l}{\textbf{all}} &
  \multicolumn{1}{l}{\textbf{NN}} &
  \multicolumn{1}{l}{\textbf{NNS}} &
  \multicolumn{1}{l}{\textbf{all}} \\ \midrule
\textbf{masc}   & 53  & 82  & 43 & 125  & 100   & 46   & 146   \\
\textbf{fem}   & 53  & 51  & 29  & 80  & 60   & 28    & 88   \\
\textbf{neut}   & 28  & 212  & 98 & 310 & 7,679 & 3,880 & 11,559 \\
\textbf{not found}  & -   & -   & -  & -   & 618   & 232   & 850   \\
\textbf{all} & 134 & 345 & 170 & 515 & 8,457 & 4,186 & 12,643 \\ \bottomrule
\end{tabular}
\caption{Composition of evaluation corpora for lexical gender detection algorithm.\\
\textbf{Note}: for \textit{Wiki1000 full}, combined predicted labels were used, because no gold labels exist for this dataset}
\label{tab:corpus_composition}
\end{table*}

\subsection{Morphological Heuristics}
Aside from the lexical database method described above, we additionally applied heuristics relating to suffix-morphology and punctuation. Morphological heuristics were applied before querying the dictionaries, while the punctuation-related heuristic was applied when a word could not be found in a dictionary.

The first heuristic was applied in order to handle gender-neutral definitions of words that carry gender-explicit markers, such as the word \textit{businessman}, which carries the masculine suffix \textit{-man}. Its definition in WordNet \citep{wordnet} is shown in \ref{example:businessman}.
\begin{enumerate}[label={(\arabic*)}, resume]
    \item \textit{businessman}: a person engaged in commercial or industrial business (especially an owner or executive)\label{example:businessman}
\end{enumerate}
Even though \textit{businessman} contains a masculine suffix, its definition is generic, most likely due to the fact that \textit{businessman} was once used for business people of all genders. However, since feminine or neutral equivalents (\textit{business woman, business person}) are widely used nowadays, the word \textit{businessman} has become gender specific and defining it generically represents an outdated, male-as-norm viewpoint \citep{fuertes-olivera2007lexical}.

We thus classified words containing the suffixes \textit{-man} and \textit{-boy} or \textit{-woman} and \textit{-girl} into masculine and feminine lexical gender, respectively. Regular expressions were used to ensure that feminine or neutral words  ending in \textit{-man} such as \textit{woman} or \textit{human}, as well as words that have the suffix \textit{-woman}, were not classified as masculine.

Another heuristic was applied in order to account for spellings that differ in punctuation, e.g. \textit{grandfather} vs. \textit{grand-father}. We check for and subsequently remove punctuation within a word if it cannot be found within a dictionary. This also applies to the cases in which non-detection is caused by a whitespace character.

\section{Data}\label{sec:data}

We used two test datasets to evaluate and run the algorithm. The first dataset, which we call \textit{gold standard} hereafter, contains nouns that have a clear lexical gender and were mainly sourced from previous research on gender bias. The second dataset contains 1,000 randomly sampled Wikipedia articles, which we used to extract gendered nouns. The following describes both datasets in detail. 

\subsection{Gold Standard}\label{ssec:gold_standard}
In order to gain insights into the performance of the dictionary-based algorithm for lexical gender retrieval, we compiled a list of words that have a nearly unambiguous lexical gender, which acts as the \textit{gold standard}.
The gold standard list was developed based on a lexical gender list by \citet{cao2020toward} with the addition of more words retrieved from online lists for learners of English\footnote{
\url{www.vocabularypage.com/2017/03/gender-specific-nouns.html}}\footnote{\url{7esl.com/gender-of-nouns/}}\footnote{\url{learnhatkey.com/what-is-gender-in-english-grammar/}
}.
Nouns retrieved from prior research and online sources were subsequently filtered for explicitness of lexical gender. For example, the pair \textit{actor/actress} would not be considered since the word \textit{actor} is nowadays used for both male and female referents. 
We moreover added neutral gender replacements for word pairs for which such an alternative exists. An example would be the triplet \textit{headmaster}-\textsc{masc}, \textit{headmistress}-\textsc{fem}, \textit{headteacher}-\textsc{neut}. The final list is comprised of 53 masculine, 53 feminine, and 28 neutral words (see Table \ref{tab:gold_standard} in the Appendix).

\begin{table*}[ht]
\centering
\begin{tabular}{@{}lcccccccc@{}}
\toprule
 &
  \multicolumn{4}{c}{\textbf{\begin{tabular}[c]{@{}c@{}}gold standard\\ (N=134)\end{tabular}}} &
  \multicolumn{4}{c}{\textbf{\begin{tabular}[c]{@{}c@{}}Wiki1000-sample\\ (N=515)\end{tabular}}} \\ \midrule
\textbf{measure}         & \textbf{P} & \textbf{R} & \textbf{F1} & \textbf{Acc}  & \textbf{P} & \textbf{R} & \textbf{F1} & \textbf{Acc}  \\ \midrule
\textbf{WordNet}         & 0.91       & 0.83       & 0.85        & \textbf{0.83} & 0.73       & 0.63       & 0.63        & \textbf{0.63}  \\
\textbf{Merriam Webster} & 0.89       & 0.77       & 0.8        & \textbf{0.77} & 0.83        & 0.82         & 0.82        & \textbf{0.82} \\
\textbf{Dictionary.com} & 0.93       & 0.87       & 0.89        & \textbf{0.87} & 0.76        & 0.61         & 0.59        & \textbf{0.61} \\
\textbf{Combined}        & 0.92       & 0.87        & 0.89        & \textbf{0.87} & 0.85       & 0.84       & 0.84        & \textbf{0.84} \\ \bottomrule
\end{tabular}
\caption{Quantitative results for lexical gender detection of gold standard and Wiki1000-sample}
\label{tab:classification_report}
\end{table*}

\subsection{Wikipedia Sample}
This research aims at providing a flexible, scalable, and high-coverage method for lexical gender detection. Therefore we additionally tested the approach on more naturalistic data, namely a random sample of 1,000 articles from English Wikipedia obtained through the \textit{wikipedia} python library\footnote{\url{https://pypi.org/project/wikipedia/}}. We will abbreviate this sample corpus as \textit{Wiki1000} hereafter.

The articles were then cleaned and tokenized into sentences using NLTK \citep{bird2009nltk} and subsequently processed with SpaCy to obtain part-of-speech (POS) tags for each word. 
All singular and plural nouns (POS-tags: NN, NNS) were then extracted and analysed for lexical gender. Nouns that contained special characters due to cleaning and tokenization errors were dropped. This method provided us with 12,643 nouns, as illustrated under Wiki1000 in Table \ref{tab:corpus_composition}. 

In order to test the performance of the algorithm, the instances of the Wiki1000 dataset needed true labels. A corpus size of 12,643 instances, however, was beyond the scope of this research to manually label. In fact, it represents the kind of corpus size that we aim to label automatically. 
We therefore filtered Wiki1000 for nouns that were labelled as either masculine or feminine by Merriam Webster Online, Dictionary.com, or WordNet. Like this, we specifically target gendered nouns and obtain a corpus similar to the gold standard corpus, but sourced from naturally occurring text. The resulting corpus of 515 nouns, which we call \textit{Wiki1000-sample}, was subsequently labelled for `true' lexical gender by members of the research team (Fleiss's $\kappa \approx 0.87$). The labels used for evaluation were determined by majority vote.
The specifications of the Wiki1000-sample dataset can be found in Table \ref{tab:corpus_composition}.

In line with previous research on gender bias in Wikipedia \cite{wagner_its_2015, falenska_assessing_2021}, which found an over-representation of male entities in the encyclopedia, Table \ref{tab:corpus_composition} shows that there are approximately 1.5 times as many mentions of distinct entities with masculine lexical gender in our 1,000-article Wikipedia sample than there of entities with feminine lexical gender.


\section{Results and Discussion}\label{sec:results}

\subsection{Quantitative analysis}

An overview of algorithm performance on the gold standard dataset and the reduced Wiki1000 sample can be found in Table \ref{tab:classification_report}. We report the weighted average of precision, recall, and F1-measure due to unbalanced classes in our test data.

As seen in Table \ref{tab:classification_report}, our best performing approach on both the gold dataset (87\% accuracy) as well as the sample of Wiki1000 (84\% accuracy) was combining labels from all three sources by majority vote. Keeping in mind that the Wiki1000 sample is approximately three times the size of the gold standard, the relative consistency in performance here indicates robustness for our approach.
It should also be noted that only querying Dictionary.com reached the same performance on the gold standard dataset (87\% accuracy) while on the Wiki1000 sample, using only Merriam Webster reached a comparable accuracy score to the combined model (82\%).

Table \ref{tab:classification_report} moreover shows that on the gold standard dataset, which was used to fine-tune our parameter values using grid search, our method reached an accuracy of 77\% or higher in each experiment configuration. Using the same parameter values for experiments on the Wiki1000 sample, only the combined approach as well as using only Merriam Webster reaches an accuracy of >77\%. When using only WordNet or Dictionary.com, the performance drops from 84\% to 63\% and 61\% accuracy, respectively. This shows that parameter configurations can be adapted to specific dictionaries and dataset sizes.

\begin{figure*}[h]

\begin{subfigure}{0.5\textwidth}
\includegraphics[width=\linewidth]{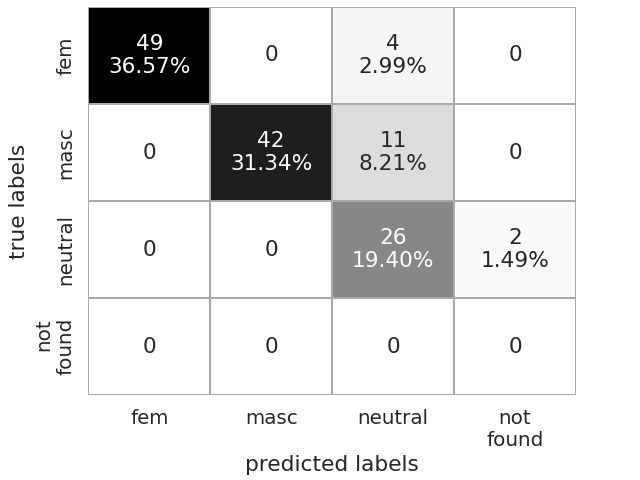} 
\caption{gold standard}
\label{fig:confmat_gold}
\end{subfigure}
\begin{subfigure}{0.5\textwidth}
\includegraphics[width=\linewidth]{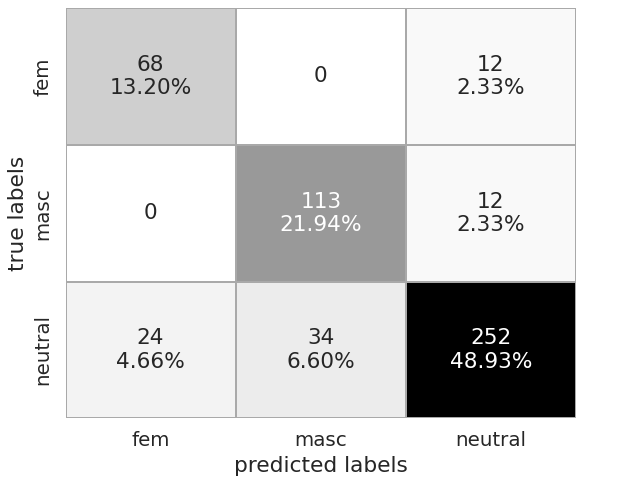}
\caption{Wiki1000-sample}
\label{fig:confmat_wiki}
\end{subfigure}

\caption{Confusion matrices for combined labels\\words that were not found in (a): \textit{single person}, \textit{child-in-law}}
\label{fig:conf_mats}
\end{figure*}

Figure \ref{fig:conf_mats} shows confusion matrices for the combined approach on both the gold standard dataset (\ref{fig:confmat_gold}) and the Wiki1000-sample (\ref{fig:confmat_wiki}). Figure \ref{fig:confmat_gold} shows that on the gold standard, the combined classifier mislabelled four feminine and 11 masculine instances as neutral, but did not mislabel any of the neutral instances as either masculine or feminine. In contrast, both these classification mistakes can be found on the Wiki1000 sample (Figure \ref{fig:confmat_wiki}). Here, the algorithm classifies more lexically neutral words as gendered than vice versa. 

Cases in which lexically neutral words are classified as gendered include words that are traditionally related to specific genders, such as \textit{bikini} or \textit{soprano}, as well as \textit{patriarchy} or \textit{testes}. It is likely that dictionary definitions reflect this traditional gender association, leading to misclassification.
Conversely, classifications of gendered words as neutral can e.g. be caused by  definitions that do not mention gender, either because of presumed knowledge (\textit{pope}) or because a lexically specific word was formerly used for all genders (\textit{landlord}). Another reason for gendered-as-neutral misclassification can be the definition of one gendered term by using another, which `cancel each other out'. For example, WordNet defines \textit{widow} as ``a \underline{woman} whose \underline{husband} is dead especially one who has not remarried'' \citep{wordnet}.

Another issue, which only occurred when testing on the gold standard dataset, concerns words that could not be found. The first is \textit{single person}, which we chose as gender-neutral alternative for \textit{bachelor}/\textit{spinster}. The fact that it was not found could be due to the term \textit{single person} being more of a composite phrase than a joined expression. Moreover, single people are often described using the adjective \textit{single} in a predicative way, such as in the sentence `He is single.', instead of `He is a single person.' 
The other word that could not be found is \textit{child-in-law}, which is the gender-neutral variant of \textit{son/daughter-in-law}. Here, the issue could be frequency of use, since \textit{child-in-law} is less established than its gender-specific variants.

\subsection{Qualitative analysis}\label{ssec:qualitative}

The following section discusses some classification errors in more detail. We focus on errors that occur due to gender-exclusive definitions in the lexical databases caused by historically close associations of words to a single gender.

In our first example, an outdated definition in WordNet \citep{wordnet} causes the misclassification of the word \textit{crew}, a neutral term, as masculine. We show the first and fourth definitions in Example \ref{example:crew} in order to illustrate how the masculine label was obtained.
\begin{enumerate}[label={(\arabic*)}, resume]
    \item \textit{crew}
    \begin{itemize}
        \item[1.] the men and women who man a vehicle (ship, aircraft, etc.)
        \item[4.] the team of men manning a racing shell 
    \end{itemize}
    \label{example:crew}
\end{enumerate}

In the first definition, the words \textit{men and women} are used to define the crew of any vehicle while in the fourth definition, which describes the crew of a racing shell (a type of rowing boat), only the word \textit{men} is used. This leads to a masculine lexical gender label, since the definitions taken together contain more masculine than feminine words. However, the fourth definition could have been worded like the first, or used the word \textit{people}, since racing shells can be crewed by people of any gender.

A similar classification error occurred for the words \textit{soprano}, \textit{menopause} and \textit{nurse}, which were all classified as feminine by the combined model, even though they have neutral lexical gender. 
These terms are all closely associated with female social gender due to anatomical and hormonal differences between sexes (\textit{soprano} and \textit{menopause}), historical biases of women performing care-work, as well as current gender distributions in certain professions (\textit{nurse}; \citeauthor{bureau_of_labor_statistics_bls_labor_2022}, \citeyear{bureau_of_labor_statistics_bls_labor_2022}).
While using gender-exclusive wording to define lexically neutral terms could inform readers of a word's traditional relation to social gender, it can also reproduce gender stereotypes and exclude those who do not identify as female but still sing in soprano voice or work as a nurse. 
Moreover, using feminine words in the definition of words like \textit{menopause} can be seen as a form of trans-exclusionary bias, since people assigned female at birth, whose body can cease to menstruate, might not identify as female.


\subsection{Limitations and Future Developments}\label{ssec:limitations}

We have selected dictionaries to obtain the lexical gender of a word, because they represent a relatively objective resource that is expected to list neutral and non-stereotypical definitions of words.
However, as shown in Section \ref{ssec:qualitative}, dictionaries are after all a human-curated resource and as such still carry human biases and outdated definitions, which in turn lead to biased or outdated results. 

We would moreover like to point out that we are explicitly working with English, which does not mark gender grammatically. In languages that mark grammatical gender, our method would most likely be obsolete, because here gender can e.g. be inferred from formal features such as morphology or agreement for most nouns \cite{corbett1991gender}.
What is more, English, as a lingua franca and the language most focused on by the NLP community \citep{bender2021dangers}, has a plethora of high-quality and high-coverage resources available. Since our method is reliant on lexical resources, adapting the method to low-resource languages could prove challenging. However, while more complex lexical resources like WordNet might not yet exist for some languages, it is likely that online dictionaries do exist. Therefore, we still believe that our method can be adapted to other notional gender languages \citep{McConnellGinet+2013+3+38}.

Another limitation of the present work concerns word sense disambiguation, since the presence of lexical gender depends on the word's sense in context.
As an example, the word \textit{colt}, can either mean a young male horse or a brand of pistol. In the sense of a male horse, the lexical gender of \textit{colt} is clearly masculine while in the sense of the pistol, it is neutral. Differences in the lexical gender of word senses can also be caused by semantic shifts, such as for the word \textit{master}, which traditionally refers to a man who is in control of e.g. servants or a household. However, in an academic context its meaning has shifted and now refers to an academic degree, or more broadly to a person of undefined gender who has reached a high level of skill in a given discipline.
Therefore, future work will integrate word sense disambiguation within the algorithm.


\section{Conclusion}\label{sec:conclusion}

We have presented a method to automatically determine the lexical gender of a given word by querying its dictionary definitions. The performance of the algorithm on a gold standard dataset of gendered nouns based on related literature, as well as a set of nouns sampled from 1,000 randomly selected Wikipedia articles, reached up to 87\% accuracy. Previous research on gender bias in NLP used manually compiled lists of gendered words for data evaluation, retrieval, manipulation, and the synthetic creation of data. In contrast, our method is scalable and has a high, dynamic coverage, which gives it a variety of applications within past and future research on gender bias in NLP. These include e.g. the assessment of gender representations in large-scale corpora, the retrieval of gendered words for which gender-neutral replacements need to be found, as well as determining whether male-centric language such as epicene \textit{he} is used in coreference resolution clusters.

\section*{Acknowledgements}
This publication has emanated from research conducted with the financial support of Science Foundation Ireland under Grant number 12/RC/2289\_P2. 
For the purpose of Open Access, the author has applied a CC BY public copyright licence to any Author Accepted Manuscript version arising from this submission.

We would like to thank Ryan O'Connor for his help in annotating the nouns in our Wikipedia corpus for lexical gender.

\bibliography{acl}
\bibliographystyle{acl_natbib}

\appendix

\section*{Appendix}\label{sec:appendix}

\begin{table*}[]
\centering
\resizebox{0.55\textwidth}{!}{%
\begin{tabular}{@{}llll@{}}
\toprule
\textbf{category}           & \textbf{masculine} & \textbf{feminine} & \textbf{neutral} \\ \midrule
\multirow{18}{*}{family}    & brother            & sister            & sibling          \\
                            & dad                & mum               &                  \\
                            & dad                & mom               &                  \\
                            & daddy              & mummy             &                  \\
                            & daddy              & mommy             &                  \\
                            & father             & mother            & parent           \\
                            & father-in-law      & mother-in-law     & parent-in-law    \\
                            & fiance             & fiancee           & betrothed        \\
                            & grandfather        & grandmother       & grandparent      \\
                            & grandson           & granddaughter     & grandchild       \\
                            & husband            & wife              & spouse           \\
                            & nephew             & niece             &                  \\
                            & son                & daughter          & child            \\
                            & son-in-law         & daughter-in-law   & child-in-law     \\
                            & step-father        & step-mother       & step-parent      \\
                            & stepfather         & stepmother        & stepparent       \\
                            & uncle              & aunt              &                  \\
                            & widower            & widow             &                  \\ \midrule
\multirow{12}{*}{misc}      & bachelor           & spinster          & single person    \\
                            & boy                & girl              & child            \\
                            & boyfriend          & girlfriend        & partner          \\
                            & gentleman          & lady              &                  \\
                            & groom              & bride             &                  \\
                            & lad                & lass              &                  \\
                            & male               & female            &                  \\
                            & man                & woman             & person           \\
                            & manservant         & maidservant       & servant          \\
                            & steward            & stewardess        & attendant        \\
                            & swain              & nymph             & spirit           \\
                            & wizard             & witch             &                  \\ \midrule
\multirow{9}{*}{occupation} & businessman        & businesswoman     & business person  \\
                            & chairman           & chairwoman        & chairperson      \\
                            & fireman            & firewoman         & fire fighter     \\
                            & headmaster         & headmistress      & head teacher     \\
                            & landlord           & landlady          & renter           \\
                            & milkman            & milkmaid          &                  \\
                            & policeman          & policewoman       & police officer   \\
                            & salesman           & saleswoman        & salesperson      \\
                            & waiter             & waitress          & server           \\ \midrule
\multirow{2}{*}{religion}   & friar              & nun               &                  \\
                            & monk               & nun               &                  \\ \midrule
\multirow{12}{*}{title}     & Mr.                & Mrs.              & Mx.              \\
                            & baron              & baroness          &                  \\
                            & count              & countess          &                  \\
                            & czar               & czarina           &                  \\
                            & duke               & duchess           &                  \\
                            & earl               & countess          &                  \\
                            & emperor            & empress           & ruler            \\
                            & king               & queen             &                  \\
                            & prince             & princess          &                  \\
                            & signor             & signora           &                  \\
                            & sir                & madam             &                  \\
                            & viscount           & viscountess       &                  \\ \bottomrule
\end{tabular}%
}
\caption{Masculine, feminine and neutral nouns of the gold standard dataset}
\label{tab:gold_standard}
\end{table*}

\end{document}